\documentclass[3p,times]{elsarticle}

\usepackage{ecrc}

\volume{00}

\firstpage{1}

\journalname{Procedia Computer Science}

\runauth{}

\jid{procs}

\jnltitlelogo{Procedia Computer Science}

\usepackage{amssymb}
\usepackage[figuresright]{rotating}
\usepackage{xcolor,soul,framed} %,caption

\colorlet{shadecolor}{yellow}
\graphicspath{../pdf/} %{../jpeg/}}
\DeclareGraphicsExtensions{.pdf,.jpeg,.png}
\usepackage[cmex10]{amsmath}
\usepackage{mathabx}
\usepackage{array}
\usepackage{mdwmath}
\usepackage{mdwtab}
\usepackage{eqparbox}
\usepackage{url}
\usepackage{amsfonts}
\usepackage{amssymb}
\usepackage{multirow}
\usepackage{pifont}
\usepackage{leftindex}

\begin{document}

\begin{frontmatter}



\dochead{}

\title{Time series forecasting for multidimensional telemetry data using GAN and BiLSTM in a Digital Twin}

 \author[jca]{João Carmo de Almeida Neto}
 \address[jca]{COPPE, Universidade Federal do Rio de Janeiro, Brazil}
 \author[cla]{Claudio Miceli de Farias}
 \address[cla]{COPPE, Universidade Federal do Rio de Janeiro, Brazil}
  \author[lea]{Leandro Santiago de Araújo} 
 \address[lea]{UFF, Universidade Federal Fluminense}
\author[leo]{Leopoldo André Dutra Lusquino Filho}
 \address[leo]{UNESP, Universidade Estadual Paulista}

\address{}

\begin{abstract}
The research related to digital twins has been increasing in recent years. Besides the mirroring of the physical word into the digital, there is the need of providing services related to the data collected and transferred to the virtual world. One of these services is the forecasting of physical part future behavior, that could lead to applicationns, like preventing harmful events or designing improvements to get better performance. One strategy used to predict any system operation it's the use of time series models like ARIMA or LSTM, and improvements were implemented using these algorithms. Recently, deep learning techniques based on generative models such as Generative Adversarial Networks (GANs) have been proposed to create time series and the use of LSTM has gained more relevance in time series forecasting, but both have limitations that restrict the forecasting results. Another issue found in the literature is the challenge of handling multivariate environments/applications in time series generation. Therefore, new methods need to be studied in order to fill these gaps and, consequently, provide better resources for creating useful digital twins. In this proposal, it's going to be studied the integration of a BiLSTM layer with a time series obtained by GAN in order to improve the forecasting of all the features provided by the dataset in terms of accuracy and, consequently, improving behaviour prediction. 
\end{abstract}


\end{frontmatter}

% \linenumbers


\section{Introduction}

Digital twins can be defined as computational models that simulate, emulate, and mirror a physical entity, which can be an object, a process, or a person \cite{barricelli2019survey}.  Digital twins applications brought possibilities like predicting failures \cite{xu2019digital,aivaliotis2019use,van2022predictive}, optimizing the configuration of a machinery \cite{jaensch2018digital, liu2022digital}, designing a process in the best optimized way \cite{liu2022digital, pan2021bim, stojanovic2018data, shiu2023digital} etc. Therefore, there are economical interests envolved, like avoiding unexpected failures in a process and improve efficiency in a industrial complex. 

\begin{figure}
  \begin{center}
  \includegraphics[width=3.5in]{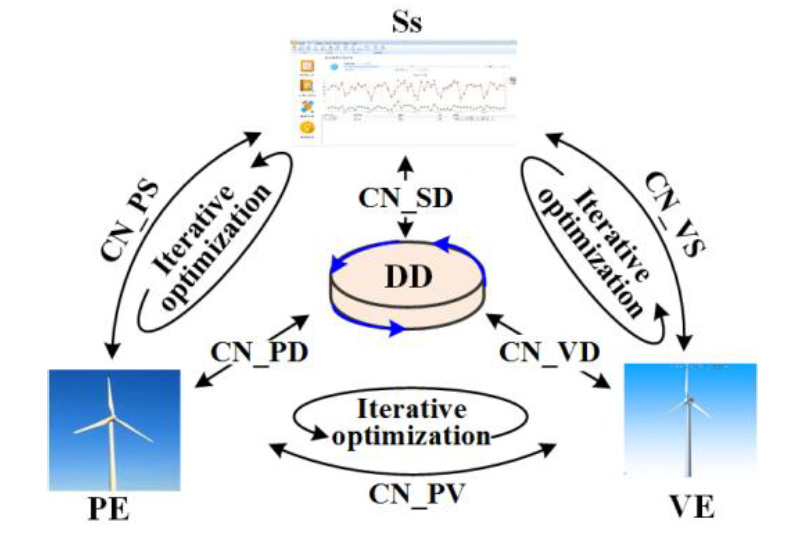}\\
  \caption{Digital Twin in Five Parts. Source \cite{tao2018digital}}\label{dt_five_parts}
  \end{center}
\end{figure}

Some definitions of Digital Twins were proposed through the years. According to \cite{tao2018digital}, Digital Twin is composed to five parts as presented in Fig. \ref{dt_five_parts}. In this figure, PE and VE represent the Physical and Virtual Entities, respectively. Ss represents the services provided by the digital twin for both PE and VE, DD represents the Digital Twin data. The most interesting part of this representation is showing the Digital Twin data as the central part of this techhnology, emphasising the importance of the data \cite{liu2023systematic}. The presented proposal is focused on the Digital Twin services (Ss).

% =======
% FIG. 01
% =======

Since most of these benefits of the services (Ss) provided by the Digital Twins are  related to predicting future behaviours of what is being monitored, the time series forecasting tends to be an important resource for predicting behaviours in different contexts \cite{huang2022time, hu2021toward, lu2021gan, wang2022improved}. To create a time series for prediction, historical data obtained from sensors in the physical system must be utilized. In this context, data-driven modelling using machine learning has become prevalent. This approach excels in extracting specific aspects of system operation, which might be challenging with model-driven or hybrid approaches. From this advantage, the presented proposal followed a data-driven method to predict multiple time series, one for each feature of the dataset.

A lot of researches have been conducted in order to forecast time series using data-driven models. Some of them use more traditional approaches like ARIMA \cite{pan2021bim,nie20223d, mwanza2024framework, kontopoulou2023review} or linear regression \cite{nottingham2001local,joseph2023time,kumar2020time}, but, more recently, Generative Adversarial Network (GAN) have been used to create time series to predict future behaviours because this deep learning architecture is able to learn data distributions \cite{quilodran2022digital,brophy2023generative}. In order to learn temporal behaviour of a time series from the dataset, LSTM has also been considered in this analysis \cite{mwanza2024framework,quilodran2022digital,anis2020optimal}.

There are data-driven studies involving GANs in the literature \cite{brophy2023generative}, but they have limitations that this proposal intends to overcome, providing a viable solution for time series forecasting in the context of a Digital Twin generation. Since Digital Twins considers the continuing operation of the environment involved, the results obtained can be naturally used to predict the its future behaviours.

In this paper, it is proposed a data-diven architecture based on GAN and BiLSTM to forecast time series data for machinery using the features extracted from sensors. In this approach, GAN is used to extract the data distribution from the dataset and the BiLSTM incorporates the temporal behaviour learned from the original dataset.

The remainder of the paper is as follows. Section II presents a short review of the literature on time series forecasting for Digital Twins. In Section III, the theoretical background of the deep learning algorithms used in our proposed model is briefly explained. The Section IV explains the proposed prediction method in details. Experimental results are provided in Section V. Section VI concludes the paper and proposes future studies.

% === II. Harmonically-Terminated Power Rectifier Analysis ========================
% =================================================================================

\section{Related Works}

Time series modelling and analysis has been historically studied and used in academic researches topics like weather modelling \cite{xu2019satellite}, finances \cite{brophy2023generative}, medicine \cite{zhu2019electrocardiogram}, construction \cite{pan2021bim}, energy consumption \cite{huang2022time} etc. Initially, more traditional approaches were used, like linear regression and ARIMA, in order to learn temporal dynamics involved but modern machine learning methods using deep learning have been considered since the data availability and computing power increased drastically in the recent years \cite{lim2021time}. As uninterrupted monitoring and data collection become more common, the need for effective temporal data analysis will increase. Therefore, it is expected that the relevance of time series forecasting data grows as well.

Time-series forecasting models predict future values of a target $y_{i,t}$ for a given entity \textit{i} at time \textit{t}. Each entity represents a logical grouping of temporal information. One-step-ahead forecasting models have the simplest algorithm considering one period into the future, and multi-step forecatisting models considers multi periods that are often beneficial in many applications \cite{lim2021time}.

Since the goal of digital twins is to mirror the real world in a virtual one, the use of time series forecastig tend to be used as an important resource, as it has been used in recent researches related to this topic \cite{quilodran2022digital, pan2021bim, yu2020prediction, anis2020optimal,lian2023anomaly,hu2021toward,kontaxoglou2021towards, mwanza2024framework,hasan2023wasserstein}. Through this method, future behaviours can be predicted and, therefore, be considered to prevent failure events and improve processes design/configuration, some important purposes to create a digital twin.

Due to the success of digital twin in manufacturing, some researches have been devoted to used it in different applications and, in terms of time series forecasting, they worth to be mentioned. In \cite{quilodran2022digital} was developed a Non-Intrusive Reduced Order digital twin to model the COVID-19 pandemic using both a method called Predictive GAN and BiLSTM. Both approaches were compared and the experiments showed that the results obtained by Predictive GAN were more precise than using BiLSTM. The novelty of this paper relies on the fact that this is was the first time a reduced order model has been applied to virus modelling.

In \cite{pan2021bim}, a digital twin was created under the integration of Building Information Modelling (BIM). Internet of Things (IoT) and data mining (DM) techniques. The research showed that the bottlenecks in the current building process can be forseen using fuzzy miner, while the number of finished tasks in the next fase could be predicted by a multivariated autoregressive integrated moving average model (ARIMAX). Consequently, tatic decision-making can realize to not only prevent possible failure in advance, but also arrange work and staffing reasonably to make the process adapt to changeable conditions. In this case, time series analysis is performed to intelligently measure and predict the successive construction progress from the future perspective.

Y. Lian et al. \cite{lian2023anomaly} proposed a digital twin-driven MTAD-GAN (Multivariate Time Series Data Anomaly Detection with GAN) oil and gas station anomaly detection method. Firstly, the operational framework consisting of digital twin model, virtual-real synchronization algorithm, anomaly detection strategy and realistic station is constructed, and an virtual-real mapping is achieved by embedding a stochastic Petri net (SPN) to describe the station-operating logic of behavior. Secondly, based on the potential correlation and complementarity among time series variables, they presente a MTAD-GAN anomaly detection method to reconstruct the error of multivariate time series to judge the abnormal samples by a given threshold.

G. Yu et al. \cite{yu2020prediction} proposed a highway tunnel pavement performance prediction approach based on a digital twin + Multiple Time Series Stacking (MTSS) for the life cycle operation and maintenance management. This proposal uses a regression approach in the time series prediction with heterogeneous stacking of extreme gradient boosting (XGBoost), the artificial neural network (ANN), random forest (RF), ridge regression, and support vector regression (SVR). It also uses a method based on multiple time series feature extraction to accurately predict the pavement performance change trend, using the highway segment as the minimum computing unit and grid search with the k-fold cross validation method to optimize hyperparameters. Lastly, this paper constructs a digital twin for pavement performance prediction to realize the real-time dynamic evolution of prediction. The method in this study is applied in the life cycle management of the Dalian highway-crossing tunnel in Shanghai, China.

Differently than other studies/methods, this proposal presents a time series generation method where the output length can be extended as needed, while keeping the dataset representing the environmental behavior updated. The method uses both GAN and BiLSTM to generate the time series and it is not constrained to a specific application or limited by data sequence length, unlike the limitations noted in GAN time series studies \cite{brophy2023generative} and those reviewed in the literature. By extending the temporal behavior of the dataset and accommodating a multivariate environment, the proposed GAN + BiLSTM approach can effectively create a Digital Twin, demonstrating the continuous operation of the modeled environment

\section{Theoretical Background}

Because the proposed architecture requires prior knowledge of GANs and LSTM/BiLSTM, an overview of these subjects is presented in the following subsections.

\subsection {Generative adversarial network (GAN)}

GAN is a generation model proposed by Goodfellow \cite{goodfellow2014generative}, which is able to learn and produce samples that follow the same distribution of the training dataset. The main elements in GAN are named discriminator (D) and generator (G), and which can be associated to various types of architectures involving machine learning models and deep neural networks. The discriminator means to find out whether a sample received is false or true. On the other hand, the objective of the generator is to deceive the discriminator by generating samples that are very similar to the original data using a random distribution as input (latent vector z). The discriminator outputs a probability indicating that a given sample is real or not. A high probability value indicates that the received sample is closer is to the real sample.

Both models D and G continue to evolve until the data generated by model G is very similar to the true data. The basic idea of GAN is the minimax game involving the generator and the discriminator.

The mathematical equations related to the discriminator D and generator G are presented as follows: 

\begin{enumerate}
    \item The loss function for discriminator D have a higher output probability of D(x) for the real sample and a lower output probability of D(G(z)) for the generated sample, as is defined in Eq. \ref{loss_discriminator}.

\begin{equation}\label{loss_discriminator}
    \begin{split}
        L_D = -\mathbb{E}_{x\sim P_{data}(x)}[logD(x)] -   \\ \mathbb{E}_{z\sim P_{z}(z)}[log(1-D(G(z)))]
    \end{split}   
\end{equation}

    \item The loss function of the generator G uses noise samples from $P_z(z)$ as input to create fake samples to deceive the discriminator D. After the GAN training process, it is expected that G capture the true data distribution. The goal of G is to maximize D(G(z)) in the loss function presented in Eq. \ref{loss_generator}.

\begin{equation}\label{loss_generator}
    L_G = \mathbb{E}_{z\sim P_z(z)}[log(1-D(G(z)))]
\end{equation}

\end{enumerate}

In both equations \ref{loss_discriminator} and \ref{loss_generator}, \textit{x} represents the real data samples, \textit{z} indicates random noise variables, G(z) represents the sample generated by G, which follows the distribution $P_{data}(x)$ of the real sample, $P_z(z)$ is the data distribution related to noise input \textit{z} and $\mathbb{E}$ indicates expectation.

The loss function for the GAN process is shown in Eq. \ref{loss_gan}.

\begin{equation}\label{loss_gan}
    \begin{split}
            \min_G \max_D L(D,G) = \mathbb{E}_{x\sim P_{data}(x)}[log(D(x))]  +  \\ \mathbb{E}_{z\sim P_z(z)}[log(1-D(G(z)))]
    \end{split}
\end{equation}

\subsection {Long Short Term Memory (LSTM) and Bidirectional Long Short Term Memory (BiLSTM)}

\begin{figure}
  \begin{center}
  \includegraphics[width=3.5in]{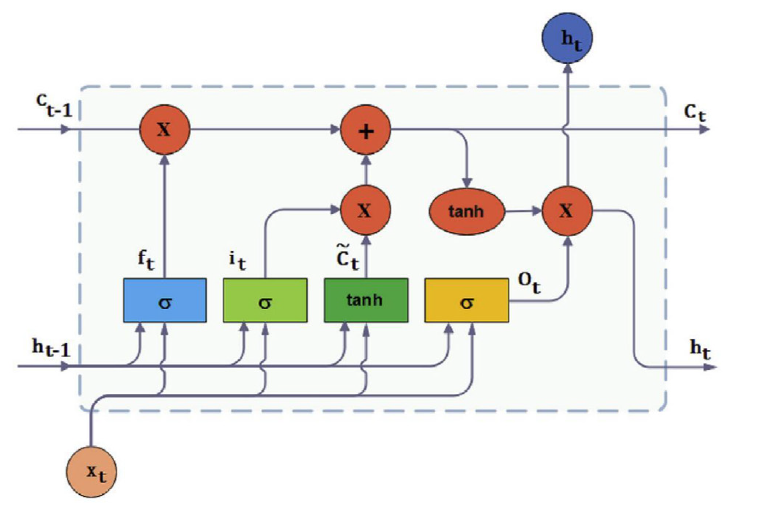}\\
  \caption{LSTM Architecture. Source \cite{huang2022time}.}\label{lstm_architecture}
  \end{center}
\end{figure}

To address the vanishing gradient problem inherent in recurrent neural networks (RNNs), a specialized type of RNN based on gates was developed, i.e., LSTM (Long Short-Term Memory). LSTM utilizes historical data to forecast the evolution of time series. LSTM utilizes historical data to forecast the evolution of time series. LSTM cells are equipped with three gate controls: the input gate ($i_t$), output gate ($o_t$) and forget gate ($f_t$) \cite{huang2022time}. The structure of LSTM is shown in Fig. \ref{lstm_architecture}.

The main calculation process of LSTM is described in the following equations:

\begin{equation}\label{input_gate_LSTM}
    i_t = \sigma(W_{ix}x_t + W_{im}m_{t-1} + W_{iC}c_{t-1} + b_i)
\end{equation}

\begin{equation}
    f_t = \sigma(W_{fx}x_t + W_{fm}m_{t-1} + W_{fc}c_{t-1} + b_f)
\end{equation}

\begin{equation}
    c_t = f_t \odot c_{t-1} + i_t \odot g(W_{cx}x_t + W_{cm}m_{t-1} + b_c)
\end{equation}

\begin{equation}
    o_t = \sigma (W_{ox}x_t + W_{om}m_{t-1} + W_{oc}c{t-1} + b_o)
\end{equation}

\begin{equation}
    m_t = o_t \odot h(c_t)
\end{equation}

\begin{equation}
    y_t = \varphi(W_{ym}m_t + b_y)
\end{equation}

where $c_t$ represents the memory cell, W and b are the weight matrices and the bias vectors, respectively, $x_t$ represents the input data at time step t, $h$ is the hidden vector, $\sigma$ is the sigmoid funtion, and $\varphi$ is the output activation function.

% =======
% FIG. 02
% =======

\begin{figure}
  \begin{center}
  \includegraphics[width=3.5in]{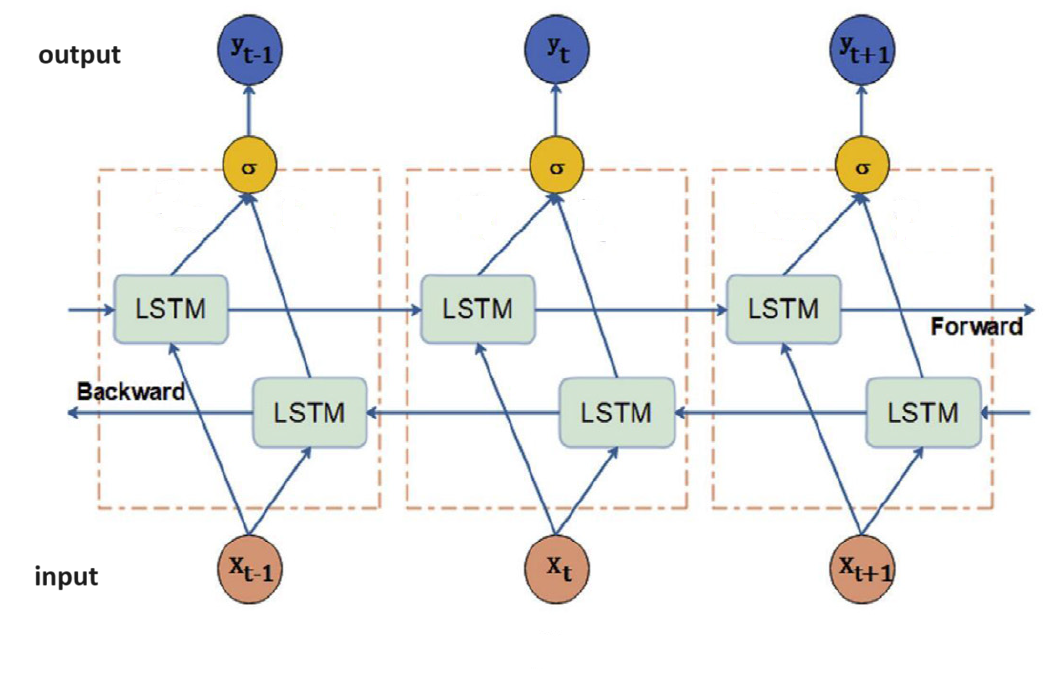}\\
  \caption{BiLSTM Architecture. Source \cite{huang2022time}.}\label{bilstm_architecture}
  \end{center}
\end{figure}

In order to process the LSTM data bidirectionally, the BiLSTM was developed using both past information in forward direction and future information in backward direction to help the cross-validation of the results and obtain high prediction accuracy compared with LSTM networks. Therefore, in order to handle the problem of “long-term dependencies” and learn the temporal correlation information from training samples, the selection of BiLSTM over LSTM seems natural because it can capture richer dependencies in the input sequence. The architecture of BiLSTM is shown in Fig. \ref{bilstm_architecture}.

% =======
% FIG. 03
% =======

There have been an increase use of BiLSTM in the last years, \cite{huang2022time,singh2023cnn,yu2020averaged,lu2021cnn,singla2022ensemble}.

\section{Proposed prediction model}

The proposed model involves sequentially combining two types of data: first, the data distribution is learned by GAN (Generative Adversarial Networks), followed by the application of the dataset's temporal characteristics through BiLSTM (Bidirectional Long Short-Term Memory). Thus, the resulting time series combines both the learned data distribution and the temporal evolution attributes of the dataset.

After training separately a GAN and a BiLSTM using a dataset divided in $n$ consecutive samples, it is necessary to use the resulting GAN Network to forecast a window of $n$ consecutive elements and use it as an input of a BiLSTM Network. The method of using GAN for prediction is called Predictive GAN \cite{quilodran2022digital, silva2021data}.

\subsection{Predictive GAN} 
\label{Predictive GAN}

The intent of the GAN generator is to produce realistic data from a randomly-generated set of latent values. In the proposed prediction process, the GAN is trained to produce $n$ time levels simultaneously. After discovering the data distribution, it is necessary to get the first $n-1$ time levels in order to predict the next time level, i.e. from known solutions ${0, 1, ..., n-1}$ the next value $n$ can be predicted.

Considering that the features of the dataset are properly reduced, obtaining Proper Orthogonal Decomposition (POD) components, and that $n-1$ values are previously known, the following steps can described in the Predictive GAN data generation:

\begin{enumerate}
    \item a latent vector $\leftindex^{0}{z}^{(n-1)}$ is randomly generated in order to start the prediction of time level $n-1$. The superscript on the left indicates the interaction counter for the optimization process;
    \item time optimisation counter is set to $n-1$;
    \item optimization iteration counter is set to 0, $l \leftarrow 0$;
    \item the output of the GAN generator is evaluated after using the current value of the latent/noise vector;
    \item the difference between the predicted and the known values is obtained;
    \item the latent vector is updated by an optimizer applying a gradient using the mean squarred error (MSE) between the predicted values and the known values. The singular values associated to each feature is used as weight during the MSE calculation;
    \item the iteration counter is incremented ($l \leftarrow l + 1 $);
    \item steps 4 to 7 are repeated until convergence is obtained;
    \item the converged latent/noise vector is used as input for the GAN generator in order to produce a window of n time steps. The value in time step $n$ is the predicted one;
    \item the predicted value in the time step $n$ is included in the known values set;
    \item go back to step 3. The process ends when time level desired is reached.
\end{enumerate}

The latent vector $z$ is randomly generated for each new time level, differently from \cite{silva2021data}, that save the latent vector recently optimized and uses it to initialise the latent variables at the next time level. This approach was chosen because showed better results in the experiments.

\begin{figure}
  \begin{center}
  \includegraphics[width= 3.5 in]{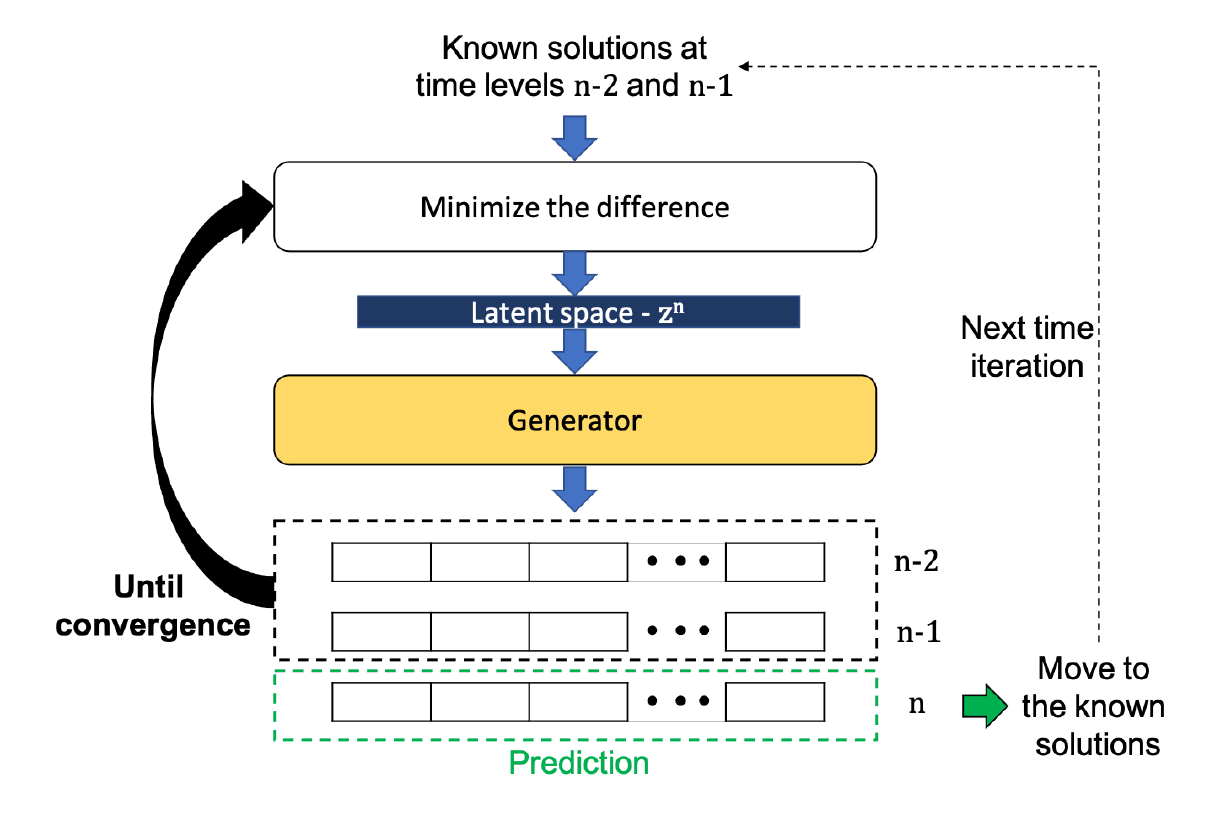}\\
  \caption{Predictive GAN iteration for a sequence of three time levels. Source \cite{silva2021data}.}\label{PredGAN-interation}
  \end{center}
\end{figure}

Fig. \ref{PredGAN-interation} illustrates how the Predictive GAN works for a generator trained to produce a sequence of 3 time steps.

% =======
% FIG. 04
% =======

\subsection{Interaction between Predictive GAN and BiLSTM}

After designing and implementing Predictive GAN, the dynamics between it and BiLSTM must be considered for the time series forecasting generation. Both time series methods are used to produce a window of $n$ time steps in order to use the output of the Predictive GAN as a input of BiLSTM. Considering this requirement, the dynamics to generate the time series forecasting can be described through the following steps:

\begin{enumerate}
    \item The trained BiLSTM forecast a window of $n$ time steps using the last window of $n$ values from the training dataset. This output is the first part of the time series forecasting proposed (window $t \leftarrow 1$);
    \item The Predictive GAN uses the last window of $n-1$ values from the training dataset to generate a time series of size $n$ according to the method explained in the section \ref{Predictive GAN}. At this point both BiLSTM and Predictive GAN have a time series forecasting with the same window size that follow the time series from the training dataset;
    \item the BiLSTM uses the last window output of size $n$ from the Predictive GAN to generate another window of $n$ forecasting time steps following the previous window output of BiLSTM ($t \leftarrow t + 1$);
    \item the Predictive GAN uses the last $n-1$ values from the previous time series window generated by the BiLSTM in order to produce another time series forecasting window of size $n$ following the previous one generated by the Predictive GAN;
    \item repeat step 3 and 4 until the desired time series forecasting size is reached.
\end{enumerate}

\begin{figure*}
  \begin{center}
  \includegraphics[width=7in]{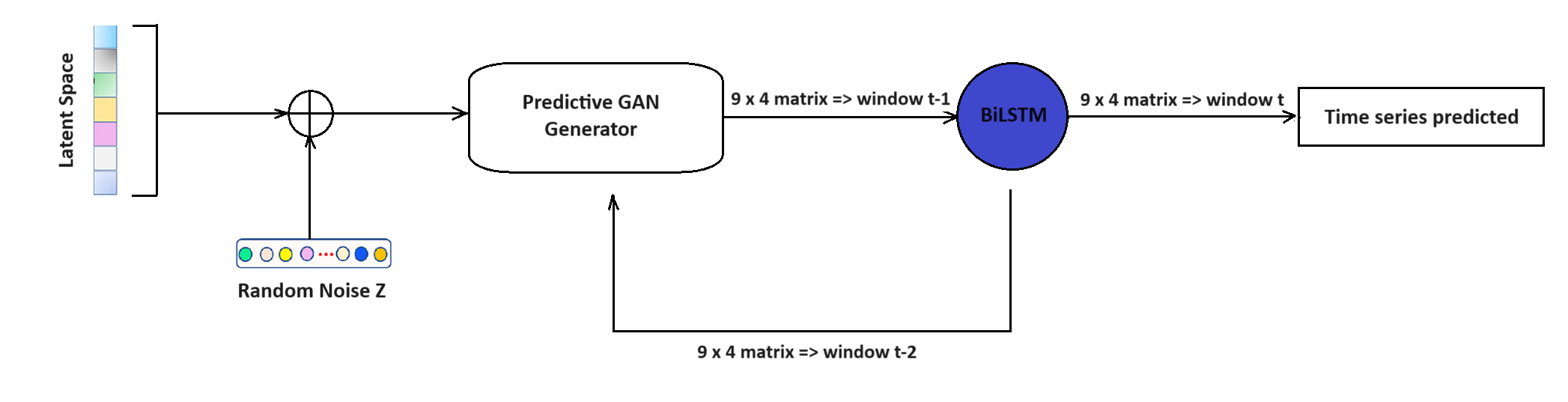}\\
  \caption{Proposed Architecture involving Predictive GAN and BiLSTM.}\label{PredGAN-BiLSTM}
  \end{center}
\end{figure*}

An example of the Predictive GAN and BiLSTM interaction can be seen in Fig. \ref{PredGAN-BiLSTM}. In this example, initially the BiLSTM have a time series forecasting until the window slot $t-1$ of size $n$. The Predictive GAN uses the $n-1$ last values from BiLSTM's window slot $t-2$ to generate its time series window slot $t-1$ of size $n$. Finally, this last window slot from the Predictive GAN's generator ($t-1$) will be used by the BiLSTM to generate the next time series forecasting window $t$. After this steps, the BiLSTM's output is a time series forecasting formed by $t$ window slots of size $n$.

% =======
% FIG. 05
% =======

As this proposal works with multivariate time series, the output of BiLSTM can be represented by an $n \times m$ matrix, where $n$ means the size of the predicted time series and $m$ represents the number of variables involved in the environment being monitored. An example can be seen in the Fig. \ref{GAN-Matrix}.

Through the integration of these two time series generation methods, it is possible to notice that it produces two time series forecasting. The first is the output of the BiLSTM after it receives the output from the Predictive GAN as an input (\textbf{PredGAN \ding{222}} \textbf{BiLSTM}). The second is the output of Predictive GAN after it uses the last $n-1$ time steps from the BiLSTM that is at the same window slot that it is, producing the next window slot of size $n$ (\textbf{BiLSTM \ding{222}} \textbf{PredGAN}). Although the first one is proposed, it is compared the precision of both time series in the following sections.

\begin{figure*}
  \begin{center}
  \includegraphics[width=4.5in]{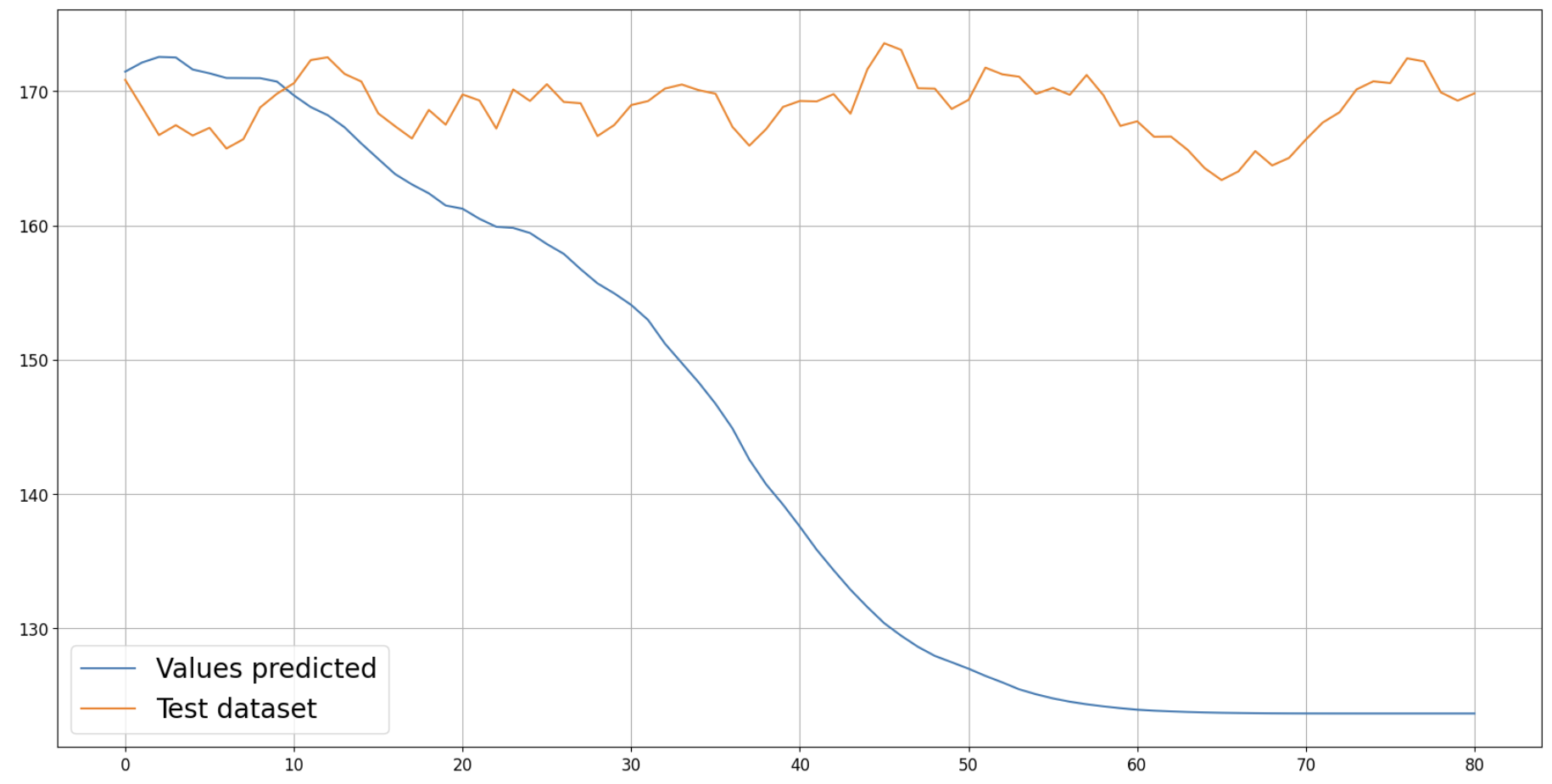}\\
  \caption{Predictive GAN used to forecast voltage values.}\label{PredGAN-Voltge}
  \end{center}
\end{figure*}

It is fair to ask why not just use Predictive GAN to make the predictions instead of integrating it with BiLSTM. As can be seen in the Fig. \ref{PredGAN-Voltge}, the predictive GAN only provides accurate temporal values up to a given amount of time steps. After that, the results become very inaccurate. Therefore, to maintain a high level of accuracy in long-term forecasting of time series, integration with BiLSTM is essential.

\section{CASE STUDY}

This part presents the details of the implemented architecture associated to the proposed prediction model. 

\subsection{Model design and related parameter selection}

\begin{figure}
  \begin{center}
  \includegraphics[width=1in]{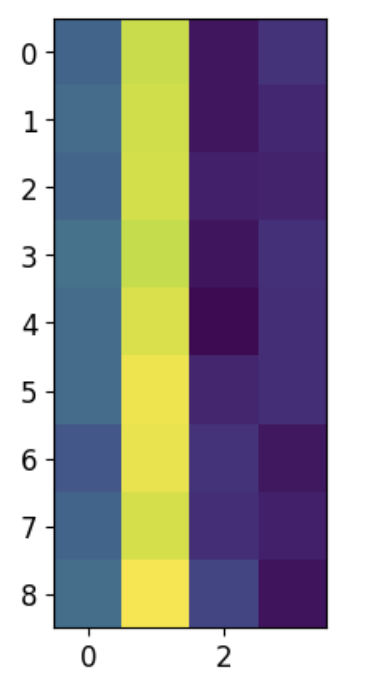}\\
  \caption{Element Matrix with size 9 x 4 used to train the GAN.}\label{GAN-Matrix}
  \end{center}
\end{figure}

The architecture involved in GAN training considers a window of nine elements in the time series to be used in the training of the generator and, consequently, the same window of nine elements to train the BiLSTM for the nine values prediction. As the dataset has four features, the time series is divided in matrices of (9x4x1) elements to be used to train the GAN (Fig. \ref{GAN-Matrix}).

% =======
% FIG. 06
% =======

Considering that GAN were designed to learn the characteristics of images and reproduce them, it is considered that the generator's output represent an image involving the values related to 9 forecasting time steps for each variable present in the input. In order to generate an image of the same size of the discriminator's input, the generator is composed by layers in the following order: 

\begin{enumerate}
    \item A Dense layer with 5120 units (5x4x256), taking a latent space noise as input, followed by batch normalization and LeakyReLU activation.;
    \item The output from the previous layer is reshaped into a matrix of 5x4x256 size;
    \item A Conv2DTranspose Layer to change the input matrix to 5x4x128 followed by a batch normalization layer and a LeakyRelu activation function;
    \item A Conv2DTranspose Layer to change the input matrix to 5x4x64  followed by a batch normalization layer and a LeakyRelu activation function;
    \item A Conv2DTranspose Layer to change the input matrix to 9x4x1  followed by a Tanh activation function.
\end{enumerate}

Once the discriminator receives an input matrix in the dimension 9x4x1, its architecture is composed by the following ordered layers:

\begin{enumerate}
    \item A Conv2D Layer to change the 9x4x1 input element into of matrix 5x4x64 size, followed by a LeakyRelu activation function and a dropout of 0.3;
    \item A Conv2D Layer to change the 5x4x64 input element into of matrix 5x4x128 size, followed by a LeakyRelu activation function and a dropout of 0.3;
    \item A Flatten Layer and a Dense Layer in order to take the 5x4x128 input into an output of one element indicating that the discriminator's input is real or fake.
\end{enumerate}

The architectures used in the generator and the discriminator was based in \cite{quilodran2022digital}.

Cross-entropy was employed to backpropagate losses during GAN training. The generator incurs a loss when the discriminator classifies its generated images as fake. Conversely, the discriminator's loss is computed as a sum of two cases: when a real image is classified as fake and when a fake image is classified as real.

The predictions obtained from predictive GAN are started by the last 8 time steps values from the training dataset in order to predict nineth value using the generator and an optimization process. The next prediction includes the last prediction value in the optimization process, now composed by 7 last dataset training values and 1 predictive GAN generated value. Therefore, after 8 iterations, the Predictive GAN works only with data from its predictions, generating the ninth prediction value.

According to the method represented in Fig. \ref{PredGAN-BiLSTM}, the predictive GAN was combined with a trained BiLSTM in the following order:

\begin{enumerate}
    \item The BiLSTM use the last 9 values from the training time series to forecast the first 9 predictions to be compared with the test dataset;
    \item The predictive GAN use the last 8 values from the training dataset to predict the first 9 time series values in the test dataset domain;
    \item The BiLSTM use the predictive GAN forecasting to predict another 9 values;
    \item The predictive GAN use the last 8 values from the first BiLSTM prediction, obtained from the training dataset, to forecast another 9 time series values;
    \item From this moment on, the BiLSTM only uses the predictive GAN forecasting to generate the 9 next values in the time series and the predictive GAN use only the last 8 BiLSTM predictions from the previous BiLSTM window to forecast 9 values.
\end{enumerate}

Therefore, this architecture use feedback strategy and the resulting time series forecasting is obtained after the BiLSTM layer.

% =======
% FIG. 07
% =======

After the architecture in Fig \ref{PredGAN-BiLSTM}, the outputs are time series for each feature of the dataset, diferently of other research that present a single value estimation \cite{huang2022time,singh2023cnn,yu2020averaged,lu2021cnn,singla2022ensemble, yang2022state, lu2021gan}. Previous research has compared GAN and BiLSTM time series individually \cite{quilodran2022digital} or focused solely on the results of one of these models \cite{yu2020averaged, elsheikh2019bidirectional}. However, there is a lack of models that combine GAN and BiLSTM to produce multivariate time series as output.

\subsection{Data source and experimental setup}

To experimentally validate the proposed approach, we used a telemetry dataset "PdM\_telemetry.csv" \cite{MSAzure-Pred_Maint}. This dataset contains four telemetry features that are used to evaluate the machines operation conditions - voltage, rotation, pressure, vibration. Although this dataset contains telemetry data from 100 machines, this paper focuses on machine 1 in order to predict the values of the features considering one particular historical data. Therefore, the original dataset was shortened to 8760 samples of this particular machine, one data per hour. Other machines in the list could have been chosen, but this one was selected only because it was the first on the list.

\begin{figure}
  \begin{center}
  \includegraphics[width=3.5in]{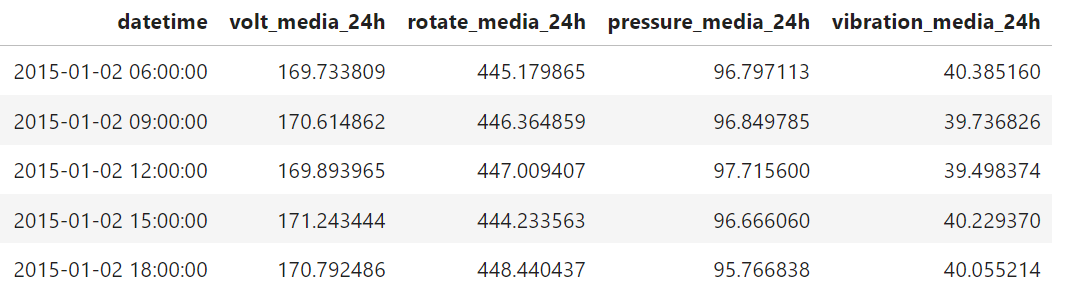}\\
  \caption{Second approach - fusion window.}\label{fusion_window}
  \end{center}
\end{figure}

In order to evaluate the best results from the dataset, removing outliers form the training dataset, this proposal uses fusion windows of three hours and a twenty four hour average of the data. Fig.\ref{fusion_window} presents a part of the complete dataset table consisting of 2914 rows, each row contains data related to the four features. The 9 x 4 matrix in Fig. \ref{GAN-Matrix} represents a sample formed by 9 consecutive rows of the complete table.

% =======
% FIG. 08
% =======

\begin{figure}
  \begin{center}
  \includegraphics[width=3.5in]{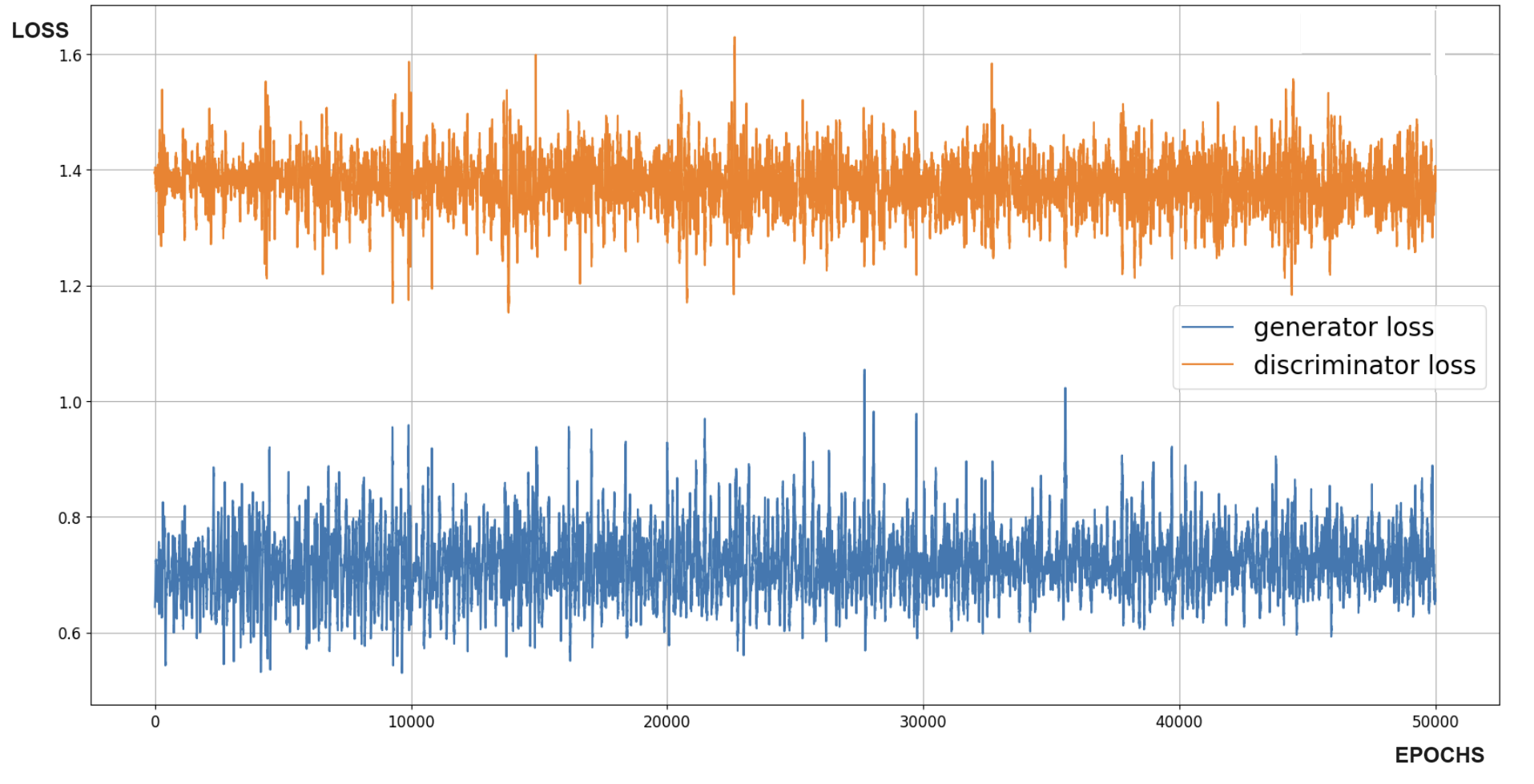}\\
  \caption{Second approach - with sliding window.}\label{gan_training_fusion_window}
  \end{center}
\end{figure}

GAN and BiLSTM models were trained using nine input samples from the time series data. The training of the GAN was completed after 50,000 epochs, as shown in Fig \ref{gan_training_fusion_window}. Each model predicted nine samples for each feature of the dataset.

% =======
% FIG. 09
% =======

\begin{figure}
  \begin{center}
  \includegraphics[width=3.5in]{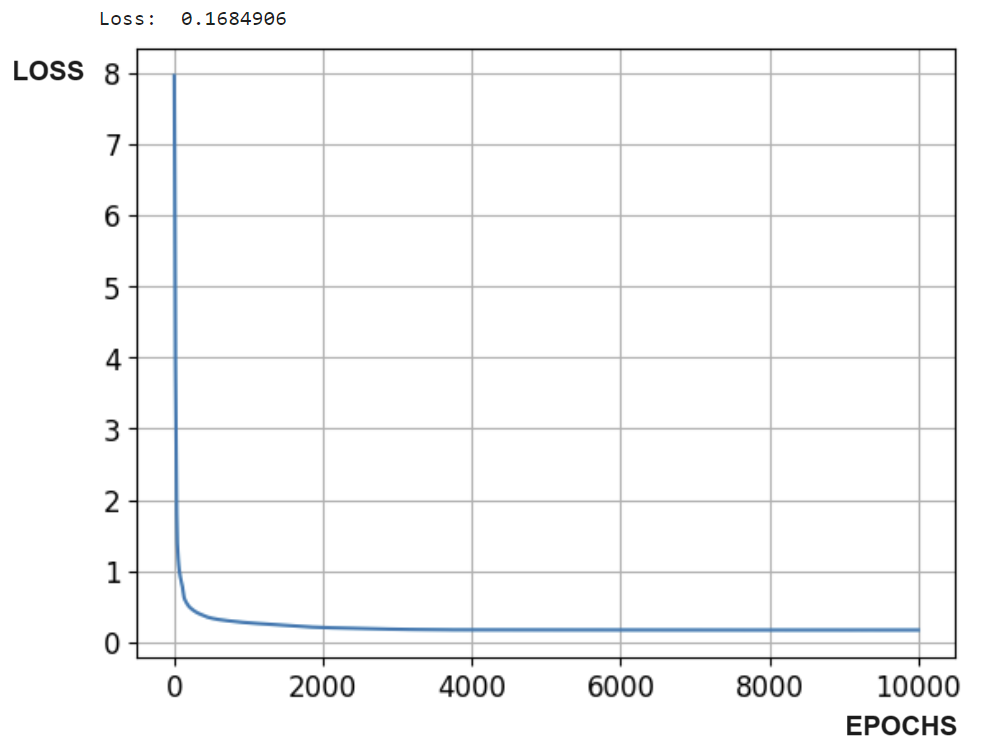}\\
  \caption{Latent Space optimization process.}\label{optimization_process}
  \end{center}
\end{figure}

The Predicted GAN output was obtained after a latent space optimization process consisting of 10,000 epochs. The evolution of the optimization process can be seen in Fig.\ref{optimization_process} and the resulting loss comparing the real value with the predicted was 0.1684906.

% =======
% FIG. 10
% =======

The dataset used in the fusion window approach was smaller than the original because of the 24 hour average and the fusion window of 3 hours, according to Fig. \ref{fusion_window}, resulting in a dataset of 2914 samples. In training,  70\% of the samples were used. 

The experiments were conducted in Python 3.11.7, Tensorflow 2.15.0 and a Jupyter notebook. The hardware plataform was a laptop using a core i9-13900HX CPU@2.2GHz, 64GB RAM, Operational System Windows 11 and a Graphic Card Nvidia RTX 4060.

\subsection{Results}

\begin{figure}
  \begin{center}
  \includegraphics[width=3.5in]{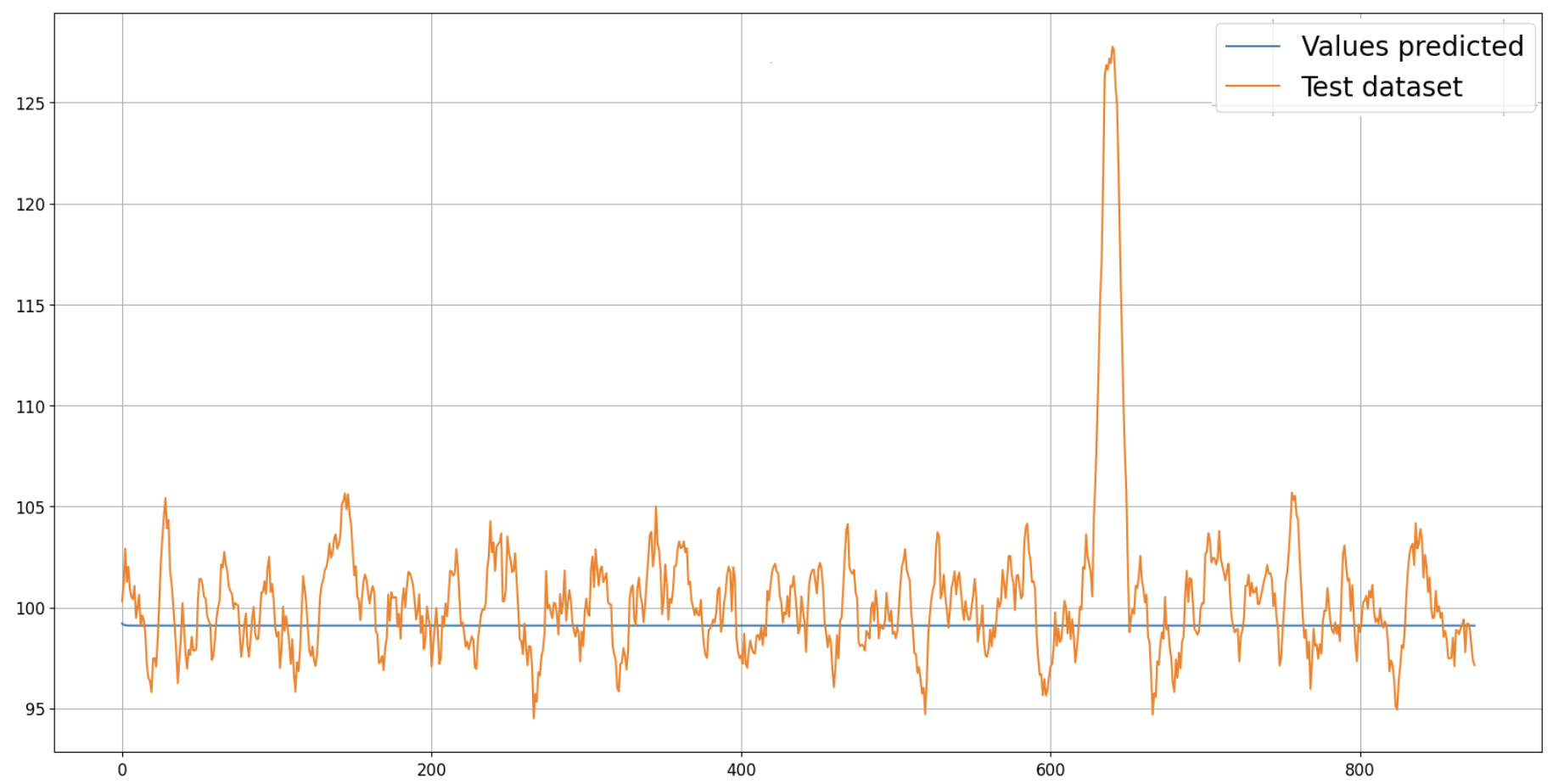}\\
  \caption{Pressure prediction using ARIMA}\label{arima_pressure}
  \end{center}
\end{figure}

Two other time series prediction models were used in benchmarking with the proposed approach. The first was ARIMA, which presented a straight line representing the average value of the time series related to the pressure values, as can be seen in Fig. \ref{arima_pressure}. The other was the Predictive GAN technique using the time series obtained from the BiLSTM as an input in order to generate a time series forecasting, opposite to the proposed method that applied the Predictive GAN output as the input of the BiLSTM. The respective values of RMSE are presented in the TABLE \ref{comparing_rmse}.

% =======
% FIG. 11
% =======

%\begin{itemize}
%    \item RMSE voltage = 5.420
%    \item RMSE rotation = 38.921
%    \item RMSE pressure = 4.183
%    \item RMSE vibration = 2.264
%\end{itemize}

%\begin{tabular}{|l|l|l|l|}
%  \hline \multicolumn{4}{|l|}{1} \\
%  \hline \multicolumn{3}{|l|}{2} & 14 \\
%  \hline 3 & \multicolumn{2}{|l|}{7} & 15 \\
%  \hline 4 & 8 & \multicolumn{2}{|l|}{12} \\
%  \hline
%\end{tabular}

\begin{table}[h]
\centering
\caption{Comparing RMSEs from diferent types of time series}
\label{comparing_rmse}
\begin{tabular}{ c|c|c|c|}
\cline{2-4}
  &\multicolumn{3}{c|} {\textbf{RMSE}} \\ [0.5ex]
  
  \hline \multicolumn{1}{ |c| } {\textbf{feature}} & \textbf{ARIMA} & \textbf{BiLSTM \ding{222}} & \textbf{PredGAN \ding{222}} \\ 
  \multicolumn{1}{ |c| } { } &  & \textbf{PredGAN} & \textbf{BiLSTM} \\
  \hline \multicolumn{1}{ |c| } {\textbf{voltage}} & 4.227 & 12.562 & 7.836 \\
  \hline \multicolumn{1}{ |c| } {\textbf{rotation}} & 19.711 & 28.486 & 22.438 \\
  \hline \multicolumn{1}{ |c| } {\textbf{pressure}} & 3.931 & 9.104 &  4.884\\
  \hline \multicolumn{1}{ |c| } {\textbf{vibration}} & 2.008 & 2.153 & 2.968 \\
  
  \hline \multicolumn{1}{ |c| } {\textbf{Average RMSE}} & 7,469 & 13,076 & 9,532 \\
  \hline
\end{tabular}
\end{table}

\begin{figure}
  \begin{center}
  \includegraphics[width=3.5in]{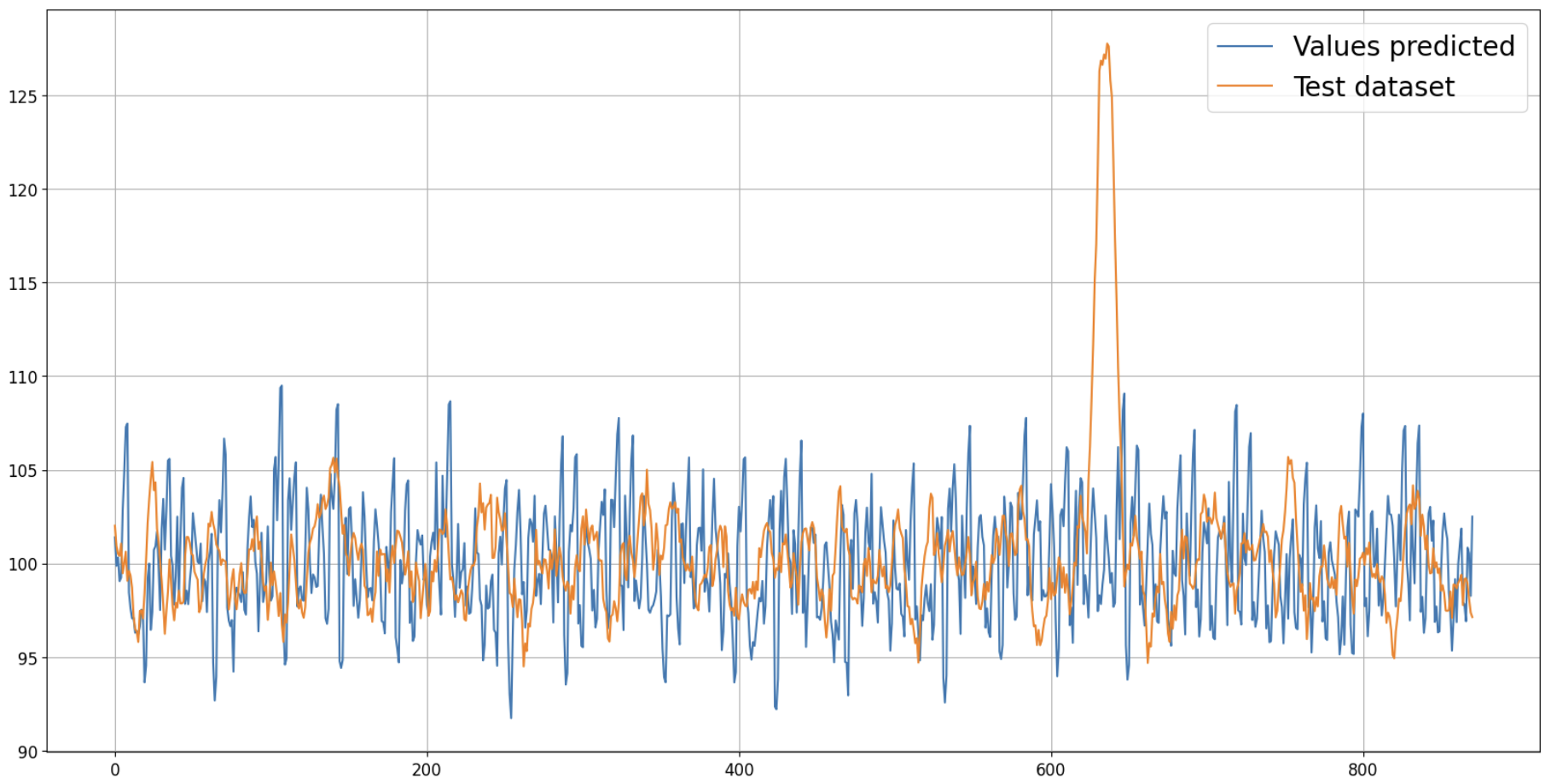}\\
  \caption{Pressure prediction using PredGAN \ding{222} BiLSTM}\label{predGAN-BiLSTM_pressure}
  \end{center}
\end{figure}

\begin{figure}
  \begin{center}
  \includegraphics[width=3.5in]{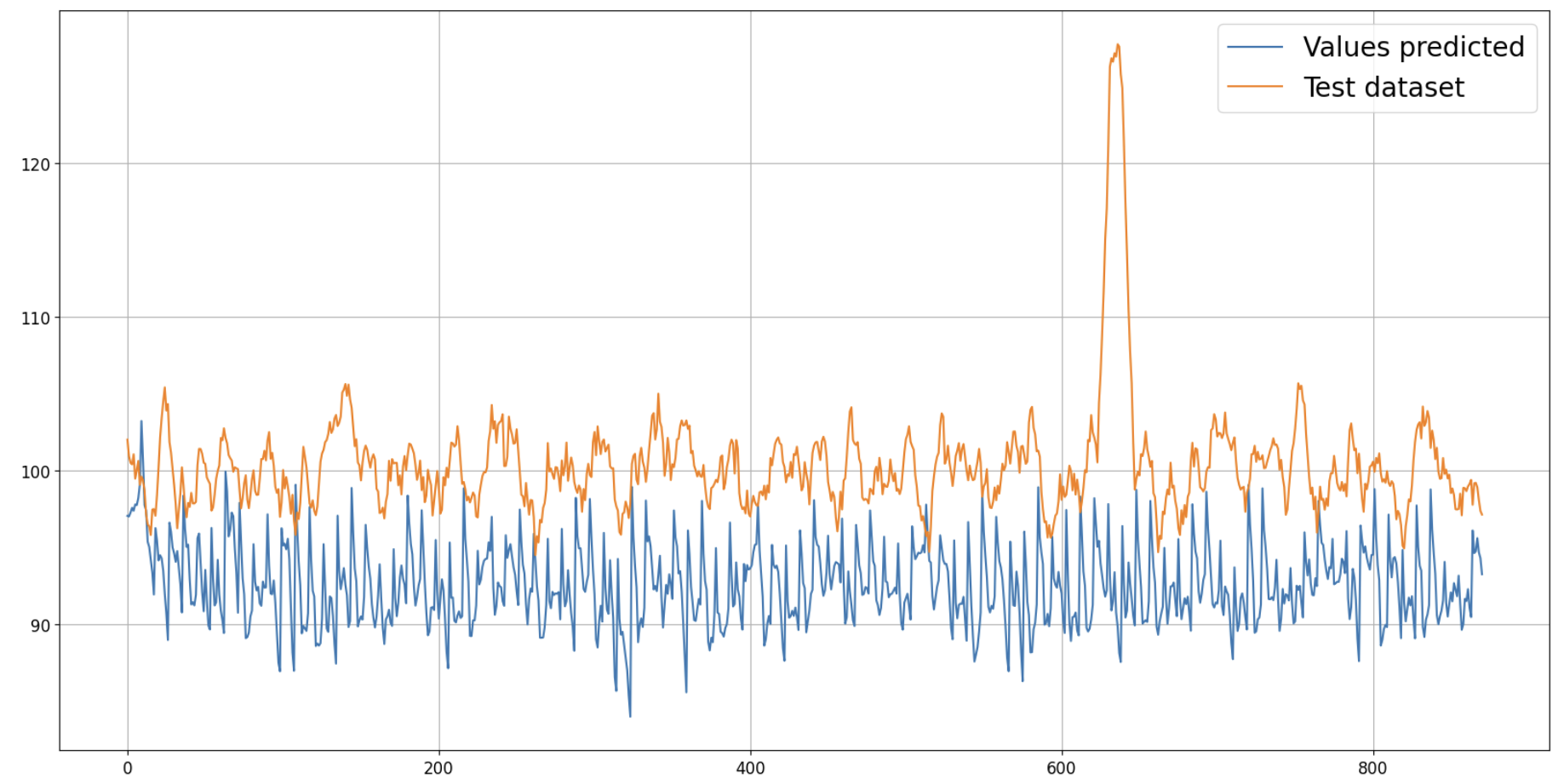}\\
  \caption{Pressure prediction using BiLSTM \ding{222} PredGAN}\label{BiLSTM-predGAN_pressure}
  \end{center}
\end{figure}

Although the average value of RMSE is smaller for the ARIMA time series, this method does not consider the variations of the features envolved, Fig. \ref{arima_pressure}, differently than the others presented, Fig. \ref{BiLSTM-predGAN_pressure} and Fig. \ref{predGAN-BiLSTM_pressure}. This proposal presents the results obtained using Predictive GAN (PredGAN) output as the BiLSTM input to generate a time series forecasting, third column, and it shows that the average RMSE is smaller than the method using the time series generated when the Predictive GAN has the BiLSTM time series as an input, second column. Therefore, the method provides better results.

% =======
% FIG. 12
% =======

% =======
% FIG. 13
% =======

\section{Conclusion}

The digital twin technology intends to provide services to its users and one of the main services is the prediction of future behaviours of the physical part being mirrored in the digital world. In this context, this paper proposes a method to generate time series forecasts for all features in the dataset. This method considers not only the data distribution of historical data through GAN but also the temporal behavior of the dataset via BiLSTM. This kind of prediction can be used to anticipate future behaviours and, therefore, anomalies. 

The experiments show that the proposed method is promising because the RMSE values do not differ significantly from the real values, indicating its capability to predict behaviours over extended time periods. It is worth mentioning that the accuracy achieved involved 70\% of the dataset samples, and higher percentages are expected to further enhance prediction accuracy.
%During the experiments, the training results using the whole dataset samples and 24 hour average samples with a fusion window of 3 hours were compared. After the experiments, it can be concluded that a simple fusion window filter using average results can improve the model accuracy, even with a smaller dataset size. It is worth to mention that the accuracy involing a 70\% samples of the dataset using fusion window provide better accuracy results than a training using 95\% samples involving the whole dataset. This result shows the positive impact of filters in the research, removing noise that compromises the forecasting time series.

In future research, its intended to explore other types of filters, such as the Kalman filter, to enhance prediction accuracy. Additionally, we aim to utilize the forecasted time series for anomaly detection, thereby facilitating calculations for the Remaining Useful Life (RUL) of the system. Another future research is to improve both the BiLSTM and the GAN hyperparameters in order to obtain better RMSE results.



%% References


\bibliographystyle{elsarticle-num}
%\bibliography{Bibliography}






% \end{thebibliography}

\end{document}